%% file: main.tex
\def\year{2021}\relax
\documentclass[letterpaper]{article} % DO NOT CHANGE THIS
\usepackage{aaai21}  % DO NOT CHANGE THIS
\usepackage{times}  % DO NOT CHANGE THIS
\usepackage{helvet} % DO NOT CHANGE THIS
\usepackage{courier}  % DO NOT CHANGE THIS
\usepackage[hyphens]{url}  % DO NOT CHANGE THIS
\usepackage{graphicx} % DO NOT CHANGE THIS
\usepackage{natbib}  % DO NOT CHANGE THIS AND DO NOT ADD ANY OPTIONS TO IT
\usepackage{caption} % DO NOT CHANGE THIS AND DO NOT ADD ANY OPTIONS TO IT
\usepackage{amsmath,amssymb}
\usepackage{subfigure}
\usepackage{algorithm}
\usepackage{algorithmicx}
\usepackage{algpseudocode}
\usepackage{mathtools}
\usepackage{makecell}
\usepackage{multirow}
\usepackage[switch]{lineno}
\usepackage{bm}

\def\etal{\emph{et al}.{}}
\def\ie{\emph{i.e.}{}}

\def\eg{\emph{e.g.}{}}

\newcommand{\PreserveBackslash}[1]{\let\temp=\\#1\let\\=\temp}
\newcolumntype{C}[1]{>{\PreserveBackslash\centering}p{#1}}
\newcolumntype{R}[1]{>{\PreserveBackslash\raggedleft}p{#1}}
\newcolumntype{L}[1]{>{\PreserveBackslash\raggedright}p{#1}}

\newcommand{\bfornot}{\normalfont}
\title{Any-Precision Deep Neural Networks}
\author{

    Haichao Yu$^{1}$, Haoxiang Li$^{2}$, Humphrey Shi$^{1,3}$, Thomas S. Huang$^{1}$, Gang Hua$^{2}$\\
}
\affiliations{
\textbf{$^{1}$UIUC, $^{2}$Wormpex AI Research, $^{3}$University of Oregon}\\
\{haichao3, hshi10, t-huang1\}@illinois.edu, haoxiang.li@bianlifeng.com, ganghua@gmail.com
}
\begin{document}
% \linenumbers
\maketitle

\begin{abstract}
  We present any-precision deep neural networks (DNNs), which are trained with a new method that allows the learned DNNs to be flexible in numerical precision during inference.
  The same model in runtime can be flexibly and directly set to different bit-widths, by truncating the least significant bits, to support dynamic speed and accuracy trade-off.
  When all layers are set to low-bits, we show that the model achieved accuracy comparable to dedicated models trained at the same precision.
  This nice property facilitates flexible deployment of deep learning models in real-world applications, where in practice trade-offs between model accuracy and runtime efficiency are often sought.
  Previous literature presents solutions to train models at each individual fixed efficiency/accuracy trade-off point. But how to produce a model flexible in runtime precision is largely unexplored. 
  When the demand of efficiency/accuracy trade-off varies from time to time or even dynamically changes in runtime, it is infeasible to re-train models accordingly, and the storage budget 
  may forbid keeping multiple models. Our proposed framework achieves this flexibility without performance degradation. More importantly, we demonstrate that this achievement is agnostic to model architectures and applicable to multiple vision tasks. Our code is released at \url{https://github.com/SHI-Labs/Any-Precision-DNNs}.
\end{abstract}

\section{Introduction}
While state-of-the-art deep learning models can achieve very high accuracy on various benchmarks, runtime cost is another crucial factor to consider in practice. 
In general, the capacity of a deep learning model is positively correlated with its complexity. As a result, accurate models mostly run slower, consume more power, 
and have larger memory footprint as well as model size. In practice, it is inevitable to balance efficiency and accuracy to get a good trade-off when deploying any deep learning models. 

\begin{figure}[t]
\centering
\hspace{-10pt}\includegraphics[width=0.8\linewidth]{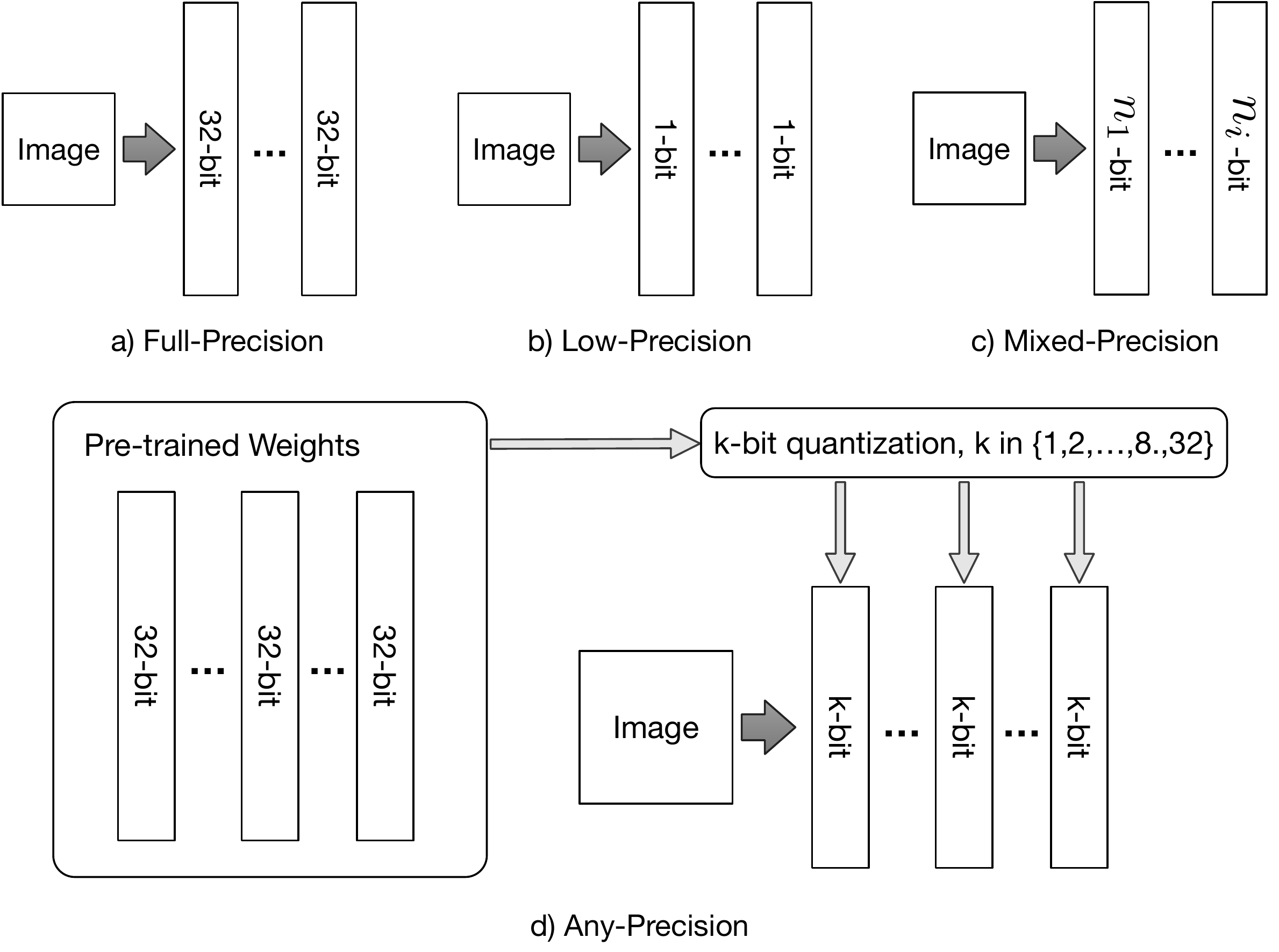}
\caption{Illustrations of deep neural networks (DNN) in different numerical precisions: a) weights and activations of typical full-precision DNN are in 32-bit floating values; b) binary DNN with 1-bit weights and activations as an example low-precision DNN; c) different layers in mixed-precision DNN can be in arbitrary bit-width and ${n_i}$ is fixed after training; d) the proposed any-precision DNN can have pre-trained weights in full-precision while in runtime the weights and activations can be quantized into arbitrary bit-width $k$.}
\label{fig.intro}
\vspace{-1em}
\end{figure}

To alleviate this issue, a number of approaches have been proposed to address it from different perspectives. We observe active researches~\cite{liu2018progressive,cai2018proxylessnas,chen2019progressive} in looking for more efficient deep neural network architectures to support practical usage~\cite{howard2019searching,yu2019universally,tan2019efficientnet}. People also consider to adaptively modify general deep learning model inference to dynamically determine the execution during the feed-forward pass to save some computation at the cost of potential accuracy drop~\cite{figurnov2017spatially,teerapittayanon2016branchynet,wu2018blockdrop,veit2018convolutional}. 

Besides these explorations, another important line of research proposes a low-level solution to use less bits to represent deep learning model and its runtime data to achieve largely reduced runtime cost. It has been shown in various literatures that full-precision is over-abundant in many applications that we can use 8-bit or even 4-bit models without obvious performance degradation. 

Some previous works went further in this direction. For example, BNN, XNOR-Net, and others~\cite{courbariaux2016binarized,rastegari2016xnor,zhou2016dorefa} are proposed to use as low as 1-bit for both the weights and activations of the deep neural networks to reduce power-usage, memory-footprint, running time, and model size. However, ultra low-precision models always observe obvious accuracy drop~\cite{courbariaux2016binarized}. While many methods have been proposed to improve accuracy of the low-precision models, so far we see no silver bullet. Stepping back from uniformly ultra-low precision models, mixed-precision models have been proposed to serve as a better trade-off~\cite{wang2019haq,dong2019hawq}. Effective ways have been found to train accurate models with some layers processing in ultra-low precision and some layers in high precision.We illustrate these different paradigms in Figure~\ref{fig.intro}.a-c. 

When we look at this spectrum of deep learning models in terms of its numerical precision, from full-precision at one end to low-precision at the other, and mixed-precision in between, we have to admit that efficiency/accuracy trade-off always exists in reality and to deploy a model in a specific application scenario we have to find the right trade-off point. Previous methods can provide a specific operating point but what if we demand flexibility as well? It would be a highly favorable property if we can dynamically change the efficiency/accuracy trade-off point given a single model. Preferably, we want to be able to adjust the model, without the need of re-training or re-calibration, to run in high accuracy mode when resources are sufficient and switch to low accuracy mode when resources are limited.

In this paper, we propose a method to train deep learning models to be flexible in numerical precision, namely any-precision deep neural networks. After training, we can freely quantize the model layers into various precision levels, without fine-tuning or calibration and without any data. We illustrate this in Figure~\ref{fig.intro}.d. When running in low-precision, full-precision or other precision levels in between, it achieves comparable accuracy to models specifically trained under the matched settings. Furthermore, given fixed computational budget, it can potentially find better operating point than one trained rigorously. 

To summarize, our contributions are:
\begin{itemize}
  \item We introduce the concept of any-precision DNN. In runtime we can quantize its layers into different bit-widths. Its accuracy changes smoothly with respect to its precision level without drastic performance degradation;
  \item We propose a novel model-agnostic method to train any-precision DNN and validate its effectiveness over multiple vision tasks, with multiple widely used benchmarks, and with multiple neural network architectures; 
  \item We demonstrate the proposed training framework trains better low-bit models with knowledge distillation.
\end{itemize}

\section{Related Work}
\paragraph{\textbf{Low-Precision Deep Neural Networks.}} 

Recent progresses in deep learning inference hardware motivate the research of using low-bit integer instead of float-point values to represent network weights and activations. 
Binarized Neural Networks~\cite{courbariaux2016binarized} and XNOR-Net~\cite{rastegari2016xnor} are early works in this direction to use only 1-bit to represent the weights and activations in DNNs.
When training these 1-bit networks, a float-point value copy of the parameters are maintained under the hood to calculate approximated gradients. Usually a {\it sign} function is used to quantize the float-point value copy to binary value in the feed-forward pass. Using only 1-bit numerical precision leads to obvious drop in accuracy in most scenarios, Zhou~\etal~\cite{zhou2016dorefa} proposed DoReFa-Net to specifically train arbitrary bitwidth in weights, activations, and gradients. Since gradients are also in low-bits, proper implementation could accelerate both the forward and backward passes. 

One of the essential problem in learning low-precision DNNs is the quantization operator. Quantization of the real-value parameters in the feed-foward pass and approximation of the gradients through the quantization operator in the backward pass heavily influence the final model accuracy. For example, the {\it sign} function adopted in Binarized NN~\cite{courbariaux2016binarized} discards the value distribution variations across layers and hurt the performance. In XNOR-Net~\cite{rastegari2016xnor}, a scaling factor is added to each layer to minimize the information loss. 
Choi~\etal~\cite{choi2018pact} proposed a parameterized clipping activation for quantization to support arbitrary bits quantization of activation. Zhang~\etal~\cite{zhang2018lq} and Jung~\etal~\cite{jung2019learning} pointed out that having an uniform quantization pattern across layers is suboptimal and propose a learnable quantizer for each layer to improve the model accuracy.

In the backward pass, most prior works use the Straight-Through Estimator (STE)~\cite{bengio2013estimating} to approximate the gradients over the quantizers. 
Cai~\etal~\cite{cai2017deep} proposed to use a half-wave gaussian quantization operator to replace the {\it sign} function for better learning efficiency and a piece-wise continuous function in the backpropagation step to alleviate the gradient mismatch issue in the prior design. Liu~\etal~\cite{liu2018bi} also attacked the gradient mismatch problem by introducing a piecewise polynomial function to approximate the {\it sign} function. Another interesting recent work from Ding~\etal~\cite{ding2019regularizing} addressed this problem by introducing a new loss function over the value distribution of layer activations.

Besides the performance gap to full-precision model, training binary networks have been reportedly to be unstable.
Tang~\etal~\cite{tang2017train} carefully analyzed the training process and concluded that using PReLu~\cite{he2015delving} activation function, a low learning rate, and the bipolar regularization on weights could lead to a more stable training process with better optimum. Zhuang~\etal~\cite{zhuang2018towards} looked at the overall training strategy and propose a progressive training process. They suggested to first train the net with quantized weights and then quantized activations, first train with high-precision and then low-precision, and jointly train the low-bit model with the full-precision one. 

A similar joint training strategy has been observed to be effective in this work as well. Since our work is along an orthogonal direction of low-precision DNNs training and design, 
our method can be complementary to train better and flexible DNNs.

\paragraph{\textbf{Post-training Quantization.}} Quantization of a pre-trained model with fine-tuning or calibration on a dataset is another related research topic in the area. 
Although methods in this area are working on a different problem from ours, we partially share the motivation to have the flexibility of quantization control in the runtime.
Without special treatment, many models collapse even in 8-bit precision in post-training quantization. One recent work from Nagel~\etal~\cite{nagel2019data} identified two issues leading to 
the large accuracy drop, the large variation in the weight ranges across channels and biased output errors due to quantization errors affecting following layers. With their method, they are able to alleviate the bias and equalize the weight ranges by rescaling and reparameterization. In this paper, the model we produced can be readily quantized into lower precision without further process. 

In the research area of deep neural networks architecture search, the slimmable neural networks by Yu~\etal~\cite{yu2018slimmable} is related to ours in terms of methodology. They presented method to 
train a single neural network with adjustable number of channels in each layer at runtime. Their exploration is limited to the search space of network architecture instead of weights.

\section{Any-Precision Deep Neural Networks}

\subsection{Overview}
Neural networks are generally constructed layer by layer. We denote input to the $i$-th layer in a neural network model as $\mathbf{x_i}$, the weights of the layer as $\mathbf{w_i}$ and the biases as $\mathbf{b_i}$.
The output from this layer can be calculated as 
\begin{equation}
  \mathbf{y_i} = \mathcal{F}(\mathbf{x_i} | \mathbf{w_i}, \mathbf{b_i}).
\end{equation}
Without loss of generality, we take one channel in a fully-connected layer as a concrete example in the following description and drop the subscript $i$ for simplicity, i.e.,
\begin{equation}
  \mathbf{y} = \mathbf{w} \cdot \mathbf{x} + b,
\end{equation}
where $\mathbf{y}, \mathbf{w} \in \mathbb{R}^{D}$ and $b$ is a scalar.

For better computation efficiency, we would like to avoid the float-value dot product of $D$-dimensional vectors.
Instead we use $N$-bit fixed-point integers to represent the weights as $\mathbf{w_Q}$ and input activations as $\mathbf{x_Q}$.
Hereafter, we assume $\mathbf{w_Q}$ and $\mathbf{x_Q}$ are stored as signed integers in its bitwise format.
Note that in some related works~\cite{courbariaux2016binarized}, elements of $\mathbf{w_Q}$ and $\mathbf{x_Q}$ could be represented as vectors of $\{-1,1\}$ and
the conversion between these two formats are trivial.
With $N$-bit integers weights and activations, as discussed in prior arts~\cite{zhou2016dorefa,rastegari2016xnor},
the computation can be accelerated by leveraging bit-wise operations ({\it and}, {\it xnor}, {\it bit-count}), or even dedicated DNN hardwares.

Early works~\cite{tang2017train} show that by adding a layer-wise real-value scaling factor $s$ could largely 
help reduce the output range variation and hence achieve better model accuracy. Since the scaling 
factor is shared across channels within the same layer, the computational cost is fractional.
Following this setting, with the quantized weights and inputs, we have 
\begin{equation}
  \mathbf{y}^{\prime} = s * (\mathbf{w_Q} \cdot \mathbf{x_Q}) + b.
  \label{eq:fc_inference}
\end{equation}
The activations $\mathbf{y}^{\prime}$ are then quantized into $N$-bit fixed-point integers as the input to the next layer.

\subsection{Inference}

We will discuss our quantization functions in details in the next section. Here we describe the runtime of a trained any-precision DNN.

\begin{figure}
\centering
\includegraphics[width=0.8\linewidth]{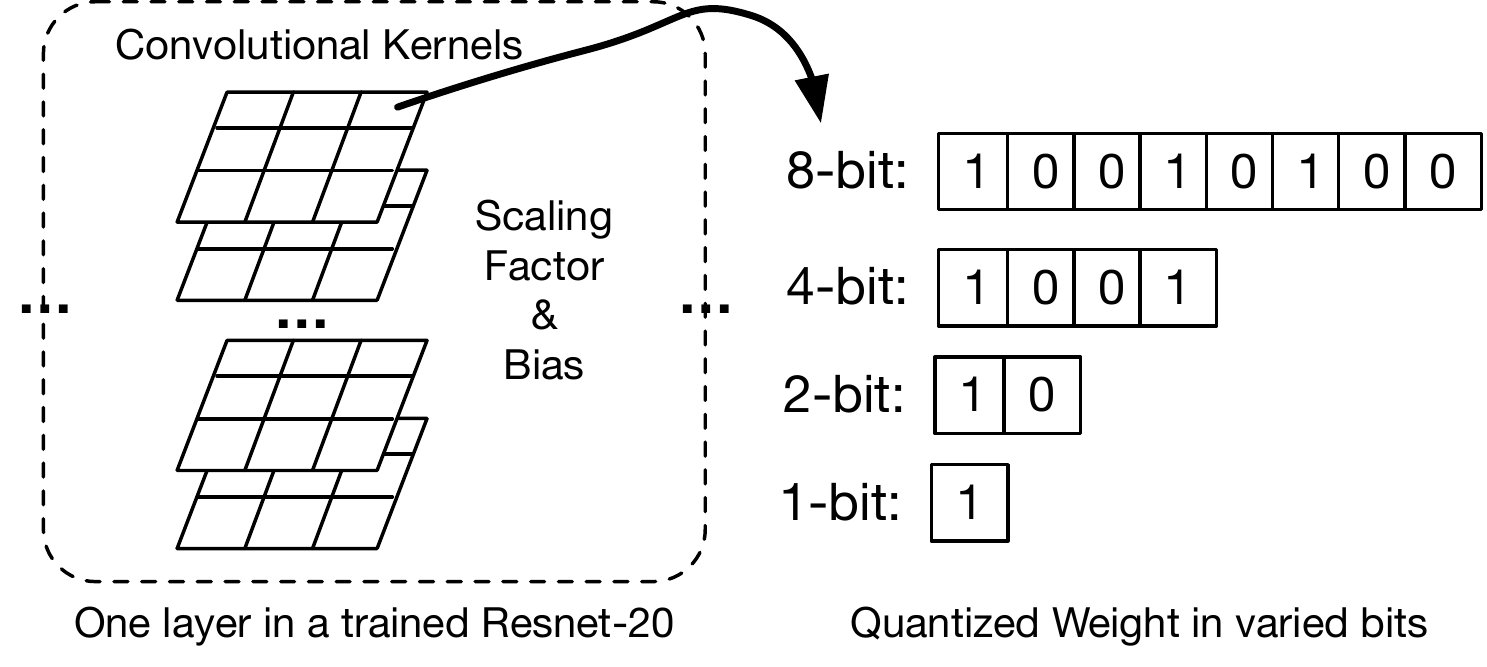}
\caption{Quantization of a kernel weight in the trained model into different precision levels: since we follow an uniform quantization pattern, when representing weight values in signed integers, the quantization can be implemented as simple bit-shift.}
\label{fig.quantization}
\vspace{-1.2em}
\end{figure}

Once training is finished, we can keep the weights at a higher precision level for storage, for example, at 8-bit. 
As shown in Figure~\ref{fig.quantization}, we can simply quantize the weights into lower bit-width by bit-shifting.
We experimentally observe that with the proposed training framework, the model accuracy changes smoothly and consistently on-par or even outperform
dedicated models trained at the same bit-width.

\subsection{Training}

A number of quantization functions have been proposed in the literature for weights and activations respectively.
Given a pre-trained DNN model, one can quantize its weights into low-bit and apply certain quantization function to activations accordingly. 
However, when the number of bits gets smaller, the accuracy quickly drops due to the rough approximation in weights and large variations in activations. 
The most widely adopted framework to obtain low-bit model is quantization-aware training. The proposed method follows the quantization-aware training framework.

We take the same fully-connected layer as an example. In training, we maintain the float-point value weights $\mathbf{w}$ for the actual layer weights $\mathbf{w_Q}$.
In the feed-forward pass, given input $\mathbf{x_Q}$, we follow Equation~\ref{eq:fc_inference} to compute the raw output $\mathbf{y}^{\prime}$.
Prior arts show the importance of the batch normalization (BN)~\cite{ioffe2015batch} layer in low-precision DNN training and we follow accordingly.
$\mathbf{y}^{\prime}$ is then passed into a BN layer and then quantized into $\mathbf{y^Q}$ as the input to the next layer.

\paragraph{\textbf{Weights.}}

We use a uniform quantization strategy similar to Zhou~\etal~\cite{zhou2016dorefa} with a scaling factor to approximate the weights. 
Given the floating point weight $\mathbf{w}$, we first apply the {\it tanh} function 
to normalize it into $[-1,1]$ and then transform it into $\mathbf{w}^\prime \in [0,1]$, i.e.,
\begin{equation}
  \mathbf{w}^\prime = \frac{\mathrm{tanh}(\mathbf{w})}{2 \mathrm{max}(|\mathrm{tanh}(\mathbf{w})|)} + 0.5.
  \label{equa:weight_normalize}
\end{equation}
We then quantize normalized value into $N$-bit integers $\mathbf{w_Q}^\prime$ and scaling factor $s$, where
\begin{equation}
% \begin{split}
  \mathbf{w_Q}^\prime = \mathrm{INT}(\mathrm{round}(\mathbf{w}^{\prime}*\mathrm{MAX_N})) \text{\ \ and\ \ }
   {s}^\prime = 1 \mathbin{/} \mathrm{MAX_N}.
  \label{eq:quantize_weights}
% \end{split}
\end{equation}
Hereafter $\mathrm{MAX_N}$ denotes the upper-bound of $N$-bit integer and $\mathrm{INT}(\cdot)$ converts a floating point value into an integer.

Finally the values are re-mapped back to approximate the range of floating point values to obtain
\begin{equation}
% \begin{split}
  \mathbf{w_Q} = 2*\mathbf{w_Q}^\prime - 1 \text{\ \ and\ \ } {s} = \mathbb{E}(|\mathbf{w}|) \mathbin{/} {\mathrm{MAX_N}},
  \label{eq:quantize_weights_final}
% \end{split}
\end{equation}
where $\mathbb{E}$ is the mean of absolute value of all floating-valued weights in the same layer. Eventually, we approximate $\mathbf{w}$ with $s * \mathbf{w_Q}$ and execute the feed-forward pass with the quantized weights as shown in Equation~\ref{eq:fc_inference}, 
the scaling factor can be applied after the dot-product of all integers vectors.

In the backward pass, gradients are computed with respect to the underlying float-value variable $\mathbf{w}$ and updates are applied to $\mathbf{w}$ as well.
In this way, the relatively unreliable and nuance signals would be accumulated gradually and hence this will stabilize the overall training process.
Since not all operations involved are nice smooth functions to support back-propagation, we use the straight through estimator (STE)~\cite{hinton2012ste} to approximate the gradients.
For example, the {\it round} operation in Equation~\ref{eq:quantize_weights} has zero derivative almost everywhere. With STE, we assign $\partial\mathrm{round}(x) \mathbin{/}  {\partial{x}} \coloneqq 1.$

\paragraph{\textbf{Activations.}}
For activation quantization in the feed-forward pass, we obtain the $N$-bit fixed-point representation by first clipping the value to be within $[0,1]$ and then
\begin{equation}
\begin{split}
    &\mathbf{y_c}^{\prime} = \mathrm{clip}(\mathbf{y}^\prime, 0, 1), \\
    &\mathbf{y_Q} = \mathrm{INT}(\mathrm{round}(\mathbf{y_c}^{\prime}*\mathrm{MAX_N})) * \frac{1}{\mathrm{MAX_N}},
\end{split}
\end{equation}
In practice, we only calculate the integer part as $\mathbf{y_Q}$ and absorb the constant scaling factor into the persistent network parameters in the next layer.

Let $L$ denote the final loss function and the gradient with respect to the activation $\mathbf{y_Q}$ is then approximated to be 
\begin{equation}
\frac{\partial{L}}{\partial{\mathbf{y_Q}}} \approx \frac{\partial{L}}{\partial{\mathbf{y_c}^\prime}}, 
\end{equation}
where
\begin{equation}
\frac{\partial{L}}{\partial{\mathbf{y_c}^\prime}} = 
  \begin{cases}
    \frac{\partial{L}}{\partial{\mathbf{y}^\prime}},& \text{if } 0\leq \mathbf{y}^\prime\leq 1, \\
    0,              & \text{otherwise}.
  \end{cases}
\end{equation}
The gradient of the ${\it round}$ function is approximated with STE to be $1$.

\begin{figure}
\centering
\vspace{-1.5em}
\includegraphics[width=0.9\linewidth]{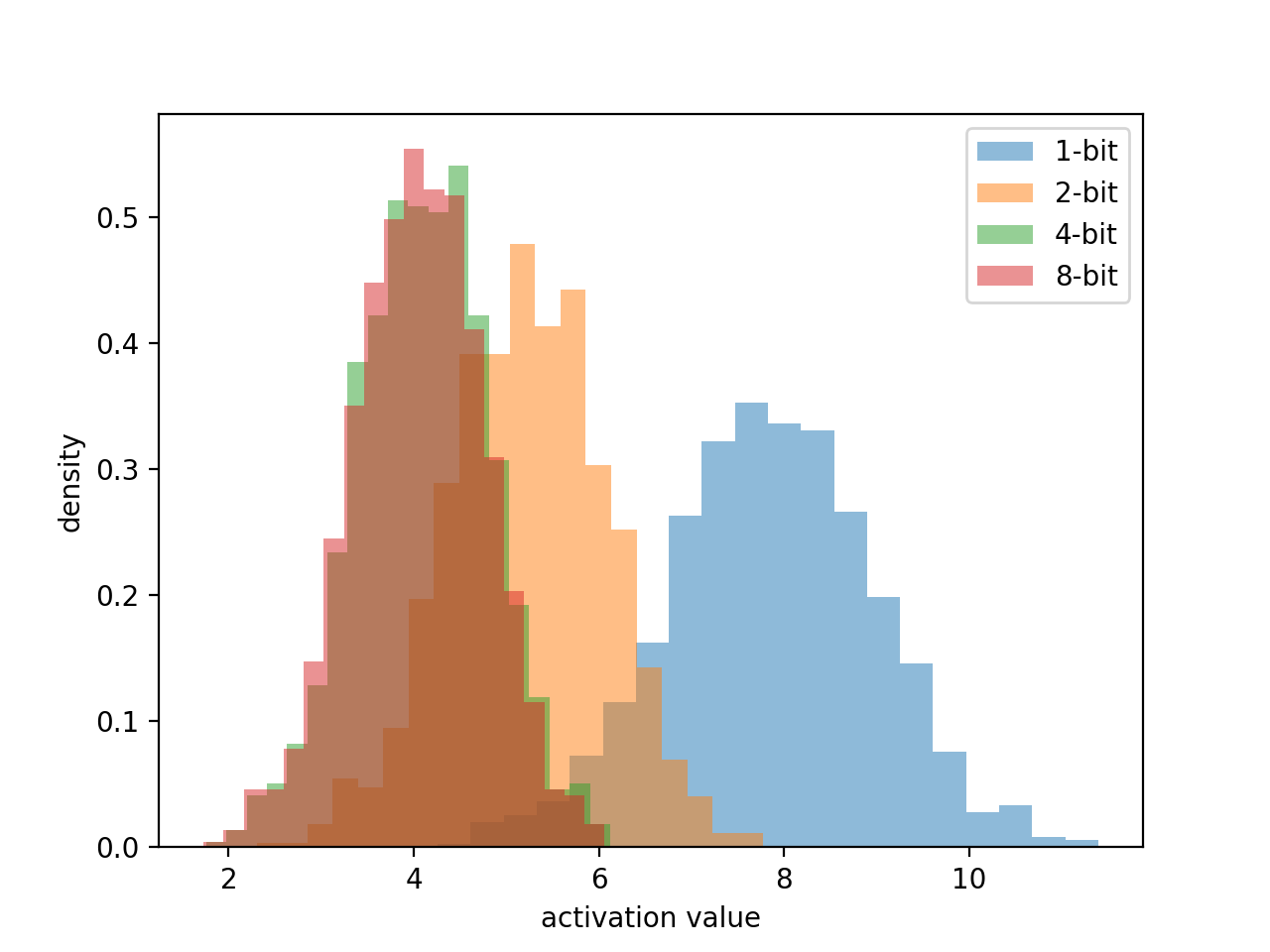}
\vspace{-0.5em}
\caption{Activation distributions under different bit-widths for weights and inputs: we randomly generate a single-channel fully-connected layer and $1000$ $16$-dimensional inputs; we then quantize the weights and inputs into ${1,2,4,8}$ bits respectively and summarize the distributions of activations under different bit-widths; as observed in the figure, the $1$-bit quantization leads to significant distribution shift compared to $8$-bit model and the discrepancy under $2$-bit is also obvious.}
\label{fig.activation}
\vspace{-1.2em}
\end{figure}

\paragraph{\textbf{Dynamic Model-wise Quantization.}}
In prior low-precision models, the bit-width $N$ is fixed during the training process. In runtime, if we alter $N$ the model accuracy drops drastically.
To encourage flexibility in the produced model, here we propose to dynamically change $N$ within the training stage to align the training and inference process.
However, the distribution of activations varies under different bit-width $N$, especially when $N$ is small (e.g., 1-bit), as shown in Figure~\ref{fig.activation}.
As a result, without special treatment, the dynamically changed $N$ creates conflicts in learning the model that it fails to converge in our experiments.

One of the widely adopted technique to adjust internal feature/activation distribution is Batch Normalization (BatchNorm)~\cite{ioffe2015batch}.
It works by normalizing layer output across batch dimension as following
\begin{equation}
    \hat{x_i} = \gamma \frac{x_i - \mu}{\sqrt{\sigma ^2 + \epsilon}} + \beta, \quad i={1..B},
\end{equation}
where $B$ is the batch size, $i$ denotes the index within current batch, $\epsilon$ is a small value added to avoid numerical issue.
$\mu$ and $\sigma^2$ are mean and variance respectively defined as 
\begin{equation}
    \mu = \sum_{i=1}^{B} x_i \text{\ \ and\ \ } \sigma^2 = \frac{1}{B} \sum_{i=1}^{B} (x_i - \mu)^2.
\end{equation}
During training, BatchNorm layer keeps calculating running averages for $\mu$ and $\sigma^2$, i.e.,
\begin{equation}
      \mu = \lambda \mu + (1-\lambda) \mu_{t} \text{\ \ and\ \ } \sigma^2 = \lambda \sigma^2 + (1-\lambda) \sigma^2_{t},
\end{equation}
where $\mu_{t}$ and $\sigma^2_{t}$ are the values before the current update, the decay rate $\lambda$ is a hyper-parameter set a-prior.
But even with the BatchNorm layer, dynamically changed $N$ will lead to failure of convergence in training due to the value distribution variations shown in the toy example in Figure~\ref{fig.activation}.

In our proposed framework, we adopted {\bf dynamically changed BatchNorm layer} to work with different $N$ in training. More specifically, assume we have a list of bit-width candidates 
$\{n_k\}^{K}_{k=1}$, we keep $|K|$ copies of BatchNorm layer parameters and internal states $\Phi^{K}_{k=1}$. When the current training iteration works with $N = n_k$, we reset the BatchNorm layers with data from $\Phi_{k}$ to use and update the corresponded copy. 

Similar technique has been adopted by Yu~\etal~\cite{yu2019universally} when dealing with varied network architectures.
Parameters of all BatchNorm layers are kept after training and used in inference. 
Note that compared with the total number of network parameters, the additional amount from BatchNorm layers is negligible.
We summarize the proposed method in Algorithm~\ref{algo:aligned}.
\begin{algorithm}[th]
  \caption{Training of the proposed any-precision DNN}
  \label{algo:aligned}
    \begin{algorithmic}[1]
      \Require{Given candidate bit-widths \(P \gets \{n_k\}^K_{k=1}\)}
    \State{Initialize the model $\mathcal{M}$ with floating-value parameters}
    \State{Initialize $K$ BatchNorm layers: $\Phi^K_{k=1}$}
    \For{\(t = 1, ..., T_{iters}\)}
        \State{Sample data batch \((x, y)\) from train set \(D_{train}\)}
        \For{\(n_p\) in \(P\)}
            \State{Set quantization bit-width $N \gets n_p$}
            \State{Feed-forward pass: \(y_{n_p} \gets \mathcal{M}(x)\)}
            % \State{Feed-forward pass: \(y^\prime_{pred} \gets \mathcal{M}(x)\)}
            \State{Set BatchNorm layers: $\mathcal{M}$.replace($\Phi_p$)}
            \State{\(L \gets L + loss(y_{n_p}, y)\)}
            %\State{\(L \gets L + loss_{distill}(y^\prime_{pred}, y_{soft})\)}
        \EndFor
        \State{Back-propagate to update network parameters}
    \EndFor
    \end{algorithmic}
\end{algorithm}
With the proposed algorithm, we can train DNN being flexible for runtime bit-width adjustment.

Another optional component in our method is {\bf adding knowledge distillation~\cite{hinton2015distilling} in training}. 
Knowledge distillation works by matching the outputs of two networks. In training a network,
we can use a more complicated model or an ensemble of models to produce soft targets by adjusting the temperature of the final softmax layer and 
then use the soft targets to guide the network learning.

In our framework, we apply this idea by generating soft targets from a high-precision model. More specifically, in each training iteration,
we first set the quantization bit-width to the highest candidate $n_K$ and run feed-forward pass to obtain soft targets $y_{soft}$.
Later, instead of accumulating cross-entropy loss for each precision candidate, we use KL divergence of the model prediction and $y_{soft}$ 
as the loss. In our experiments, we observe that in general knowledge distillation leads to better performance at low-bit precision levels.

\section{Experiments}
We first validate our method with several network architectures and datasets on image classification task. These networks include a $8$-layer CNN (named Model C in~\cite{zhou2016dorefa}), AlexNet~\cite{krizhevsky2012imagenet}, MobileNetV2~\cite{sandler2018mobilenetv2}, Resnet-8, Resnet-18, Resnet-20 and Resnet-50~\cite{he2016deep}. The datasets include Cifar-10~\cite{krizhevsky2009learning}, Street View House Numbers (SVHN)~\cite{netzer2011reading}, and ImageNet~\cite{imagenet_cvpr09}.
We also evaluate our method on the image segmentation task to demonstrate its generalization.

\subsection{Implementation Details}
We implement the whole framework in PyTorch~\cite{paszke2017automatic}. 
On Cifar-10, we train AlexNet, MobileNetV2 and Resnet-20 models for 400 epochs with initial learning rate $0.001$ and decayed by $0.1$ at epochs $\{150,250,350\}$. 
On SVHN, the 8-layer CNN named CNN-8 and Resnet-8 models are trained for 100 epochs with initial learning rate $0.001$ and decayed by $0.1$ at epochs $\{50,75,90\}$. 
We combine the training and extra training data on SVHN as our training dataset. 
All models on Cifar-10 and SVHN are optimized with the Adam optimizer~\cite{kingma2014adam} without weight decay.
On ImageNet, we train Resnet-18 and Resnet-50 dedicated models for 120 epochs with initial learning rate $0.1$ decayed by $0.1$ at epochs $\{30,60,85,95,105\}$ with SGD optimizer. For any-precision model, we trained 80 epochs with initial learning $0.3$ decayed by $0.1$ at epochs $\{45,60,70\}$.

For all models, following Zhou~\etal~\cite{zhou2016dorefa} we keep first and last layer real-valued. 
In training, we train the networks with bit-width candidates $\{1,2,4,8,32\}$. Note that when the bit-width is set to $32$, 
it is a full-precision model that we use floating-valued weights and activations.
In testing, we evaluate the model at each bit-width in the list respectively. 
By default, we recursively add knowledge distillation (KD) in training. Concretely, we use the full-precision model to 
get soft targets as supervision for the 8-bit model, the ones from the 8-bit model for the 4-bit model, so on and so forth.

\begin{table*}[ht]
    \begin{center}
    \renewcommand{\arraystretch}{1.0}
    \caption{Comparison of the proposed any-precision DNN to dedicated models: the proposed method achieved the strong baseline accuracy in most cases, even occasionally outperforms the baselines in low-precision. We hypothesize that the gain is mainly from the knowledge distillation from high-precision models in training. Dedi.: Dedicated models.} 
    \label{tab:compare2baseline}
    \begin{tabular}{c|c|cc|cc|cc|cc|cc}
    \hline
    \multirow{2}{*}{Datasets} & \multirow{2}{*}{Models} & \multicolumn{2}{c|}{1 bit} & \multicolumn{2}{c|}{2 bit} & \multicolumn{2}{c|}{4 bit} & \multicolumn{2}{c|}{8 bit} & \multicolumn{2}{c}{FP32} \\
    \cline{3-12}
    & & Dedi. & Ours & Dedi. & Ours & Dedi. & Ours & Dedi. & Ours & Dedi. & Ours \\
    \hline
    \multirow{3}{*}{Cifar-10} &
    Resnet-20   &92.07&\bfornot{92.15}&93.55&\bfornot{93.97}&93.71&\bfornot{93.95}&93.66&\bfornot{93.80}&\bfornot{94.08}&93.98\\
                              &AlexNet     &
    92.56&\bfornot{93.00}&94.06&\bfornot{94.08}&94.02&\bfornot{94.26}&93.82&\bfornot{94.24}&93.74&\bfornot{94.22}\\
                              & MobileNetV2 &
    \bfornot{80.17}&80.06&89.69&\bfornot{90.74}&92.01&\bfornot{92.27}&\bfornot{92.44}&92.32&\bfornot{94.01}&93.84 \\
    \hline
    \multirow{2}{*}{SVHN}     & Resnet-8    &
    \bfornot{92.94}&91.65&\bfornot{95.91}&94.78&95.15&\bfornot{95.46}&94.64&\bfornot{95.39}&94.60&\bfornot{95.36}\\
                              & CNN-8       &
    \bfornot{90.94}&88.21&\bfornot{96.45}&94.94&\bfornot{97.04}&96.19&\bfornot{97.04}&96.22&\bfornot{97.10}&96.29\\
    \hline
    \multirow{2}{*}{ImageNet} & Resnet-18   &
    \bfornot{55.06}&54.62&63.65&\bfornot{64.19}&\bfornot{68.15}&67.96&\bfornot{68.48}&68.04&\bfornot{69.27}&68.16\\
                              & Resnet-50   &
    61.08&\bfornot{63.18}&69.44&\bfornot{73.24}&71.24&\bfornot{74.75}&74.71&\bfornot{74.91}&\bfornot{75.95}&74.96\\
    \hline
    \end{tabular}
    \end{center}
\vspace{-1em}
\end{table*}

\begin{table}[ht]
    \centering
    % \small
    \caption{Resnet-50 model size comparison (in MegaByte): the proposed model only brings little overhead upon size of a FP32 model but achieves the flexibility what could take 5 independent models instead. Ded: Dedicated models.}
    \label{tab:compareparams}
    \begin{tabular}{c|C{0.8cm}|C{0.52cm}|C{0.52cm}|C{0.52cm}|C{0.52cm}|C{0.72cm}|C{0.8cm}}
    \hline
                         & Shared & 1 & 2 & 4 & 8 & 32 & Total \\
    \hline
    Ded            & 0      & 22.3 & 24.9 & 30.1 & 40.4 & 102.4 & 220.1 \\
    Ours                 & 102.0 & 0.4  & 0.4  & 0.4  & 0.4  & 0.4   & 104.0 \\
    \hline
    \end{tabular}
\end{table}

\subsection{Comparison to Dedicated Models}
We compare our method to very competitive baseline models at each precision level. For each bit-width we tested, we dedicatedly train a low-precision model following the same training pipeline with fixed bit-width for weights and activations. We compare the accuracy we obtained from our dedicated low-bit models to other recent works in this field to make sure the baseline models are strong. For example, on Cifar-10, our 1-bit baseline achieved an accuracy of $92.07\%$ while the recent work from Ding~\etal~\cite{ding2019regularizing} reported $89.90\%$.

Our results are summarized in Table~\ref{tab:compare2baseline}. We observed that binarizing depth-wise convolution layers in MobileNetV2 may lead to training divergence, which is verified by Hai~\etal~\cite{phan2020mobinet}.
Therefore, we replace depth-wise convolutions with group convolutions in MobileNetV2 models.
As shown in the table, on all three datasets, the proposed any-precision DNN generally achieved comparable performance to the competitive dedicated models. At the same time,
our model is more compact. As shown in Table~\ref{tab:compareparams}, compared with five individual models, our any-precision model (104MB) saved more than 50\% parameters (220MB). 

\begin{table}[htb]
    \begin{center}
    \caption{Comparison to other post-training quantization methods: All models are Resnet-50 trained on ImageNet. When bit-width drops from their original training setting, our method consistently outperform them. BS: Bit-Shifting from FP32 model. BC: BatchNorm Calibration from FP32 model.}
    \label{tab:compare2postimagenet}
    \begin{tabular}{C{2.2cm}|C{0.72cm}|C{0.72cm}|C{0.75cm}|C{0.72cm}|C{0.72cm}}
        \hline
        {\scriptsize Models \textbackslash~Runtime bit}  & 1 & 2 & 4 & 8 & FP32\\ 
        \hline
        BS  & 0.100 & 0.136 & 42.70 & 74.68 & \bfornot{75.95} \\
        % \hline
        BS+BC & 0.106 & 0.352 & 57.12 & 73.73 & \bfornot{75.95} \\
        % \hline
        ACIQ & 0.116 & 0.324 & 71.64\,\footnotemark & \bfornot{75.83} & \bfornot{75.95} \\
        % \hline
        Ours  & \bfornot{63.18} & \bfornot{73.24} & \bfornot{74.75} & 74.91 & 74.96 \\
        \hline
    \end{tabular}
    \end{center}
    \vspace{-1.5em}
\end{table}

\subsection{Post-Training Quantization Methods~\label{sec:post}}
We compare our method with three alternative post-training quantization methods. We experiment with Resnet-50 on ImageNet.

The first naive baseline directly quantizes dedicated models with bit-shifting. In other words, to obtain an $(n-1)$-bit model from a trained $n$-bit model, as what is done with any-precision DNN shown in Figure~\ref{fig.quantization}, we simply drop the least-significant bit of all weights. With no surprise, this strategy fails dramatically on challenging large-scale benchmark as shown in Table~\ref{tab:compare2postimagenet}.

The second strategy follows the same bit-shifting to drop bit with an added BatchNorm calibration process. In the calibration process, BatchNorm statistics will be re-calculated by feed-forwarding a number of training samples. As shown in 
Table~\ref{tab:compare2postimagenet}, the BatchNorm calibration helped in 4-bit but still failed in 1,2-bit settings. 

% for tab:compare2postimagenet
\footnotetext{We used the paper's official code \url{https://github.com/submission2019/cnn-quantization}, but we cannot reproduce the claimed accuracy 73.8\% on 4-bit model.}

With the proposed method, we can leverage this post-training calibration technique to fill-in the gaps of training candidate bit-width list, i.e., after training  for 1,2,4,8,32-bits precision levels, we can further calibrate the model under the remaining 3,5,6,7-bit settings to get the missed copies of BatchNorm layer parameters. So that, in runtime, we can freely choose any precision level from $1$ to $8$ bits.

In addition, we compare with a recently proposed method ACIQ~\cite{banner2019post}, which introduced analytical weight clipping, adaptive bit allocation and bias correction for post-training quantization. As shown in Table~\ref{tab:compare2postimagenet}, our method achieved much better accuracy under low bit-width settings.

\begin{figure}[t]
\centering
\vspace{-1em}
\includegraphics[width=.49\textwidth]{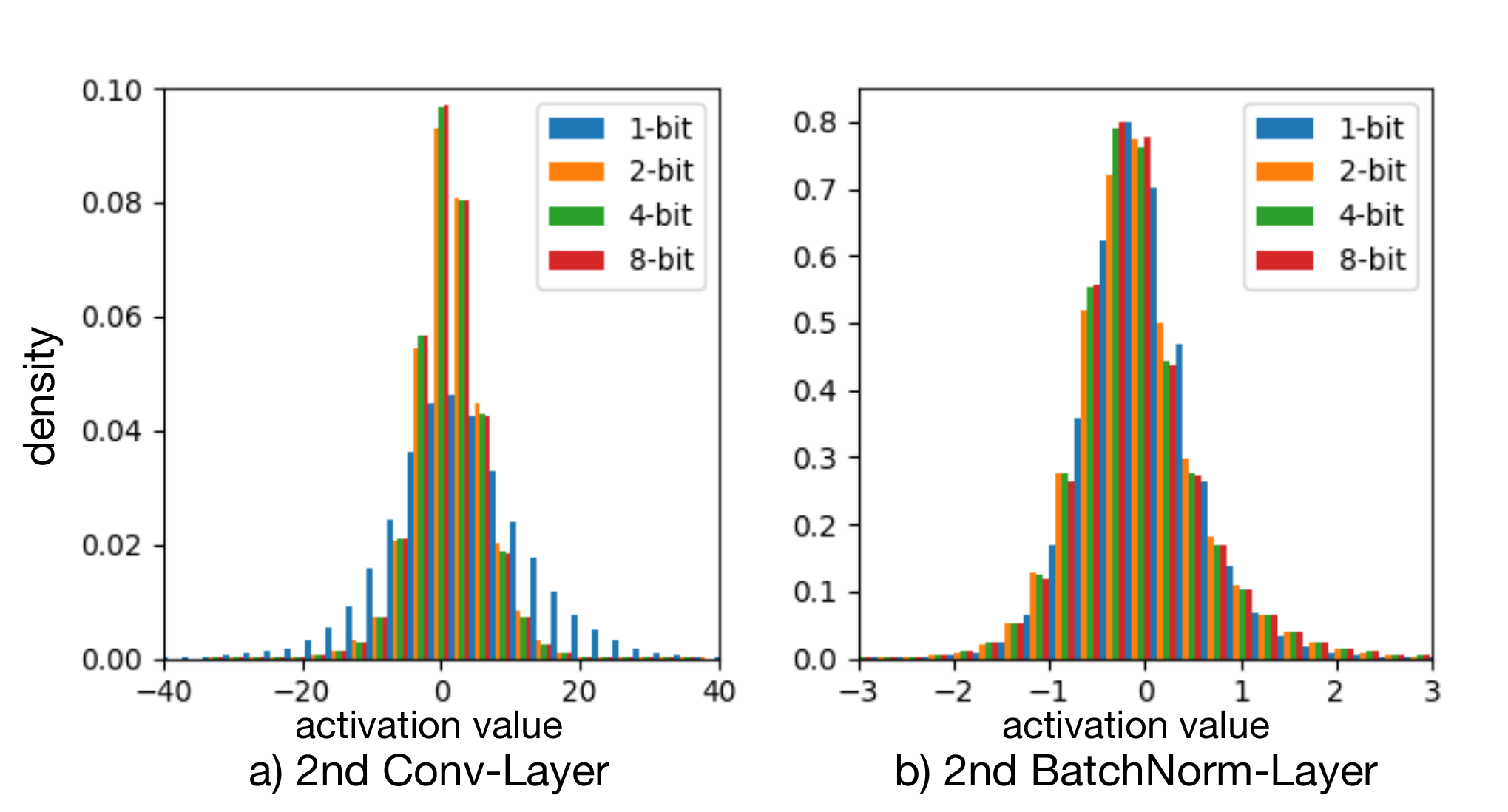}
\caption{Activation value distributions of several layers in an any-precision AlexNet: low-bit quantization leads to value distribution different from others after convolutional layers but accordingly changed BatchNorm layer could rectify the mis-match.}
\label{fig:activation_dist_3layers}
\end{figure}

\subsection{Dynamically Changed BatchNorm Layers}
\begin{table}[t]
    \centering
    \small
    \caption{\small Classification accuracy of Resnet-20 with different bit-width combinations in training on Cifar-10.}
    \label{tab:abla}
      \begin{tabular}{C{1.05cm}|C{0.45cm}C{0.45cm}C{0.45cm}C{0.45cm}C{0.45cm}C{0.45cm}C{0.45cm}C{0.45cm}}
        \hline
        {\tiny Train \textbackslash~Test} & 1 & 2 & 3 & 4 & 5 & 6 & 7 & 8 \\
        \hline
        1,2,4,8 & 91.80 & 93.48 & 93.22 & 93.70 & 93.58 & 93.52 & 93.53 & 93.61 \\
        \hline
        1,8     & 91.95 & 89.57 & 92.97 & 93.23 & 93.35 & 93.34 & 93.36 & 93.39 \\
        \hline
        2,8     & 10.00 & 93.58 & 93.50 & 93.83 & 93.91 & 93.90 & 94.00 & 93.93 \\
        \hline
        4,8     & 10.00 & 72.78 & 93.22 & 93.80 & 93.75 & 93.79 & 93.81 & 93.79 \\
        \hline
    \end{tabular}
    \vspace{-1.5em}
\end{table}

To understand how the dynamically changed BatchNorm layers help in our framework, we visualize the activation value distributions of several layers of an any-precision AlexNet. More specifically, we look at how activation value distribution changes from the 2nd convolutional layers and the BatchNorm layers after them when the runtime precision level is set to $1$,$2$,$4$,$8$-bit respectively. As shown in Figure~\ref{fig:activation_dist_3layers}, when running at $1$-bit precision, the activation distribution is obviously off from others after the convolutional layers; the followed BatchNorm layer rectifies the distributions; then the next convolutional layer would create this distribution variation again. It is very clear that by keeping multiple copies of the BatchNorm layer parameters for different bit-widths, we can minimize input variations to the convolutional layers and hence have the same set of convolutional layer parameters to support any-precision in runtime.

\subsection{Ablation Studies}
\paragraph{\textbf{Candidate bit-width List.}}
We study how the candidate bit-width list used during training the any-precision DNN influences the testing performance on other bit-widths. 

Table~\ref{tab:abla} shows testing accuracy of models trained under different bit-width combinations. We observe that training with more candidate bit-width generally leads to better generalization
to the others and the candidate bit-width list is better to cover the extreme cases in runtime. For example, the 1,8-bits combination performs more stable across different runtime bit-width
compared with 2,8-bits and 4,8-bits combination. Since better coverage in training takes longer for the model to converge, this observation can guide the bit-width selection under limited training resources.

\begin{table}[t]
    \centering
    \caption{Impact of knowledge distillation on classification accuracy.}
    \label{tab:abla_kd}
    % \footnotesize
    \begin{tabular}{C{1.2cm}|C{0.8cm}|C{0.7cm}|C{0.7cm}|C{0.7cm}|C{0.7cm}|C{0.7cm}}
        \hline
        \multicolumn{2}{c|}{Models \textbackslash~Bits} & 1 & 2 & 4 & 8 & FP32 \\
        \hline
        \multirow{2}{*}{\scriptsize\shortstack{Resnet-20\\on Cifar-10}} & {\scriptsize w/o KD} & 91.67 & 93.72 & 93.83 & \bfornot{93.94} & 93.82 \\
        & {\scriptsize KD} & \bfornot{92.15} & \bfornot{93.97} & \bfornot{93.95} & 93.80 & \bfornot{93.98} \\
        \hline
        \multirow{2}{*}{\scriptsize\shortstack{Resnet-50\\on ImageNet}} & {\scriptsize w/o KD} & 62.42 & 72.88 & 74.43 & 74.68 & \bfornot{74.98} \\
        & {\scriptsize KD} & \bfornot{63.18} & \bfornot{73.24} & \bfornot{74.75} & \bfornot{74.91} & 74.96 \\
        \hline
    \end{tabular}
    \vspace{-1em}
\end{table}

\vspace{-1em}
\paragraph{\textbf{Knowledge Distillation.}}
We study the influence of knowledge distillation during any-precision training. The first case is {\it w/o KD}, \ie, no KD is used as shown in Algorithm~\ref{algo:aligned}). The second case is {\it KD}, \ie, the highest bit-width outputs are supervised by groundtruth, and then every other bit-width outputs are supervised by the output logits from the nearest superior bit-width.

In Table~\ref{tab:abla_kd}, we observed on both Resnet-20 and Resnet-50, KD outperforms the dedicated models in low-bit settings. The hypothesis is that for lower-bit models, soft logits from higher-bit models instead of groundtruth labels regularize the training better.

\begin{table}[htb]
    \centering
    % \footnotesize
    \caption{Segmentation performance comparison on Pascal VOC 2012 segmentation dataset under mIoU and top-1 accuracy.}
    \label{tab:seg}
    \begin{tabular}{c|c|c|c|c|c|c}
    \hline
    \multicolumn{2}{c|}{Models \textbackslash~Bits} & 1 & 2 & 4 & 8 & FP32 \\
    \hline
    \multirow{2}{*}{Dedicated} & mIoU & 59.5  & 71.5  & 74.1  & \bfornot{75.6}  & \bfornot{76.2} \\
    & Acc. & 89.7 & 93.5 & 94.2 & \bfornot{94.5} & \bfornot{94.6}\\
    \hline
    \multirow{2}{*}{Ours} & mIoU & \bfornot{61.3}  & \bfornot{72.1} & \bfornot{74.3}  & 75.4  & 75.9 \\
    & Acc. & \bfornot{90.6} & \bfornot{93.7} & \bfornot{94.3} & 94.4 & \bfornot{94.6}\\
    \hline
    \end{tabular}
    \vspace{-1.2em}
\end{table}

\subsection{Application to Semantic Segmentation}
To demonstrate generalization of our method to other tasks, we apply the any-precision scheme to semantic segmentation. We train Deeplab V3~\cite{chen2017rethinking} with Resnet-50 on Pascal VOC 2012 segmentation dataset~\cite{everingham2015pascal} with SBD dataset~\cite{BharathICCV2011} as groundtruth augmentation. We use the publicly available PyTorch training codebase~\footnote{https://github.com/kazuto1011/deeplab-pytorch}. All models are pretrained from the corresponding classification models on ImageNet. We do not use any post-processing for segmentation.

We summarize our results in Table~\ref{tab:seg} under two evaluation metrics mIoU and top-1 accuracy. Our low-bit models perform better than the dedicated counterparts. In 8,32-bit settings, our models achieve comparable results. Similar to the results on the classification tasks, we attribute this improvement to the joint optimization and knowledge distillation.

\section{Discussion}
We study how the multiple bit-width joint training influences parameter learning. With STE, the parameter being updated in training is essentially the FP32 weight $w$. The gradient on $w$ comes from losses from different bit-widths. The necessary condition for the joint training to work is that the gradients from different bit-widths are consistent with each other. 

Inspired by \cite{zhao2018modulation}, we analyze gradient consistency between bit-widths by computing Update Compliance Average (UCA), which is defined as the average of cosine similarity of gradients from two bit-widths over multiple training steps. During joint training, we observed that different bit-widths share very large UCA ($[0.9, 1]$), indicating consistent gradient directions and thus training convergence. We also found that neighboring bit-widths share larger UCA than the others, \eg, UCA between 1 and 2 bits ($0.929$) are larger than 1 and the other bits (0.909 between 1 and 32 bits). This motivated us to employ recursive knowledge distillation in joint training. 
Refer Appendix~\ref{app:uca} for more details.

As indicated by UCA, we also observed a small gradient direction gap. We think this makes our model naturally more robust to adversarial attacks than a single dedicated model as a by-product from joint training. By switching to different bit-widths, any-precision model under one bit-width can show robustness to adversarial attacks targeting at another bit-width. We experimented with FGSM attacking method~\cite{ian2014explaining}.  Any-precision models show improved defensive performance compared with dedicated ones, \eg, classification accuracy of FP32 model is improved by more than 10\%.
See Appendix~\ref{app:robust} for more details.
We believe our any-precision scheme can serve as an add-on to other defensive methods against adversarial attacks.

\section{Conclusion}
In this paper, we introduce any-precision DNN to address the practical efficiency/accuracy trade-off dilemma from a new perspective. Instead of seeking for a better operating point, we enable runtime adjustment of model precision-level to support flexible efficiency/accuracy trade-off without additional storage or computation cost. The model can be stored at 8-bit or higher and run in lower bit-width specified at runtime. Our flexible model achieves comparable accuracy to dedicatedly trained low-precision models and surpasses other post-training quantization methods. 

\bibliography{references.bib}

\clearpage
\input{supp.tex}

\end{document}

%% file: supp.tex
\appendix
\section*{Appendices}

\section{Discussion}
\subsection{Update Compliance Average Analysis}
\label{app:uca}
Inspired by \cite{zhao2018modulation}, we analyze gradient consistency between bit-widths by computing Update Compliance Average (UCA). Considering gradients of a layer as a vector $\bm{g}$, UCA is defined as
\begin{equation}
    UCA = \frac{1}{NL} \sum_{n=1}^{N} \sum_{l=1}^{L} \frac{\langle \bm{g}_{l,n}, \bm{g}^\prime_{l,n}\rangle}{\|\bm{g}_{l,n}\| \|\bm{g}^\prime_{l,n}\|}.
\end{equation}
$N$ and $L$ are training steps and network layer number. $\bm{g}_{l,n}$ and $\bm{g}^\prime_{l,n}$ are gradient vector from two different bits in layer $l$ at training step $n$. $\langle\cdot,\cdot\rangle$ is vector inner product. In Table~\ref{tab:uca}, we show UCA of Resnet-20~\cite{he2016deep} trained on Cifar-10~\cite{krizhevsky2009learning} over 50 epochs. In the table, all UCAs are in $[0.9, 1]$, which indicates consistent gradient directions and thus training convergence. In addition, neighboring bit-widths share larger UCA than the others. For example, UCA between 1 and 2 bits are larger than 1 and the other bits, and UCA between 32 and 8 bits are larger than 32 and ther other bits.

\begin{table}[ht]
    \centering
    \caption{UCA between different bit-widths}
    \label{tab:uca}
    \begin{tabular}{c|ccccc}
        \hline
        Bit \textbackslash~Bit & 1 & 2 & 4 & 8 & 32 \\
        \hline
        1 & 1.000 &  0.929 &  0.910 &  0.909 &  0.909 \\
        2 & - &  1.000 &  0.955 &  0.954 &  0.955 \\
        4 & - &  - &  1.000 &  0.963 &  0.963 \\
        8 & - &  - &  - &  1.000 &  0.964 \\
        32 & - &  - &  - &  - &  1.000 \\
        \hline
    \end{tabular}
\end{table}

\subsection{Robustness to Adversarial Attacks}
\label{app:robust}
We experiment with FGSM attacking method with step size $\epsilon=0.007$~\cite{ian2014explaining}. In Table.~\ref{tab:robustness}, we show classification accuracy of dedicated and any-precision models on adversarial examples. When running with the bit-width that is targeted by adversarial attacks, our model gets similar or better classification accuracy compared with the dedicated counterpart. When our model switches between different bit-widths when being attacked, it shows better defensive ability.

\begin{table}[ht]
    \centering
    \caption{Classification accuracy on adversarial examples. Ded: Dedicated model. $B_A$: bit being attacked. $B_D$: bit used for defense. Dedicated model is attacked and makes defense under the same bit-width.}
    \label{tab:robustness}
    \begin{tabular}{C{0.7cm}|c|C{0.72cm}C{0.72cm}C{0.72cm}C{0.72cm}C{0.72cm}}
    \hline
    & $B_A$\textbackslash$B_D$ & 1 & 2 & 4 & 8 & FP32 \\
    \hline
    Ded    & - & 75.09 & 72.80 & 70.45 & 70.23 & 70.00  \\
    \hline
    \multirow{5}{*}{Ours} & 1 & 75.65 & 83.87 & 84.78 & 84.87 & 84.25 \\
                          & 2 & 79.45 & 74.91 & 78.18 & 78.77 & 78.55 \\
                          & 4 & 81.22 & 78.18 & 72.93 & 74.27 & 75.85 \\
                          & 8 & 81.58 & 78.65 & 74.28 & 73.32 & 76.06 \\
                          & FP32 & 81.95 & 80.46 & 78.26 & 78.33 & 75.41 \\
    \hline
    \end{tabular}
\end{table}

%% file: main.bbl
\begin{thebibliography}{45}
\providecommand{\natexlab}[1]{#1}
\providecommand{\url}[1]{\texttt{#1}}
\providecommand{\urlprefix}{URL }
\expandafter\ifx\csname urlstyle\endcsname\relax
  \providecommand{\doi}[1]{doi:\discretionary{}{}{}#1}\else
  \providecommand{\doi}{doi:\discretionary{}{}{}\begingroup
  \urlstyle{rm}\Url}\fi

\bibitem[{Banner, Nahshan, and Soudry(2019)}]{banner2019post}
Banner, R.; Nahshan, Y.; and Soudry, D. 2019.
\newblock Post training 4-bit quantization of convolutional networks for
  rapid-deployment.
\newblock In \emph{Advances in Neural Information Processing Systems},
  7948--7956.

\bibitem[{Bengio, L{\'e}onard, and Courville(2013)}]{bengio2013estimating}
Bengio, Y.; L{\'e}onard, N.; and Courville, A. 2013.
\newblock Estimating or propagating gradients through stochastic neurons for
  conditional computation.
\newblock \emph{arXiv preprint arXiv:1308.3432} .

\bibitem[{Cai, Zhu, and Han(2018)}]{cai2018proxylessnas}
Cai, H.; Zhu, L.; and Han, S. 2018.
\newblock Proxylessnas: Direct neural architecture search on target task and
  hardware.
\newblock \emph{arXiv preprint arXiv:1812.00332} .

\bibitem[{Cai et~al.(2017)Cai, He, Sun, and Vasconcelos}]{cai2017deep}
Cai, Z.; He, X.; Sun, J.; and Vasconcelos, N. 2017.
\newblock Deep learning with low precision by half-wave gaussian quantization.
\newblock In \emph{Proceedings of the IEEE Conference on Computer Vision and
  Pattern Recognition}, 5918--5926.

\bibitem[{Chen et~al.(2017)Chen, Papandreou, Schroff, and
  Adam}]{chen2017rethinking}
Chen, L.-C.; Papandreou, G.; Schroff, F.; and Adam, H. 2017.
\newblock Rethinking atrous convolution for semantic image segmentation.
\newblock \emph{arXiv preprint arXiv:1706.05587} .

\bibitem[{Chen et~al.(2019)Chen, Xie, Wu, and Tian}]{chen2019progressive}
Chen, X.; Xie, L.; Wu, J.; and Tian, Q. 2019.
\newblock Progressive Differentiable Architecture Search: Bridging the Depth
  Gap between Search and Evaluation.
\newblock \emph{arXiv preprint arXiv:1904.12760} .

\bibitem[{Choi et~al.(2018)Choi, Wang, Venkataramani, Chuang, Srinivasan, and
  Gopalakrishnan}]{choi2018pact}
Choi, J.; Wang, Z.; Venkataramani, S.; Chuang, P. I.-J.; Srinivasan, V.; and
  Gopalakrishnan, K. 2018.
\newblock Pact: Parameterized clipping activation for quantized neural
  networks.
\newblock \emph{arXiv preprint arXiv:1805.06085} .

\bibitem[{Courbariaux et~al.(2016)Courbariaux, Hubara, Soudry, El-Yaniv, and
  Bengio}]{courbariaux2016binarized}
Courbariaux, M.; Hubara, I.; Soudry, D.; El-Yaniv, R.; and Bengio, Y. 2016.
\newblock Binarized neural networks: Training deep neural networks with weights
  and activations constrained to+ 1 or-1.
\newblock \emph{arXiv preprint arXiv:1602.02830} .

\bibitem[{Deng et~al.(2009)Deng, Dong, Socher, Li, Li, and
  Fei-Fei}]{imagenet_cvpr09}
Deng, J.; Dong, W.; Socher, R.; Li, L.-J.; Li, K.; and Fei-Fei, L. 2009.
\newblock {ImageNet: A Large-Scale Hierarchical Image Database}.
\newblock In \emph{CVPR09}.

\bibitem[{Ding et~al.(2019)Ding, Chin, Liu, and
  Marculescu}]{ding2019regularizing}
Ding, R.; Chin, T.-W.; Liu, Z.; and Marculescu, D. 2019.
\newblock Regularizing activation distribution for training binarized deep
  networks.
\newblock In \emph{Proceedings of the IEEE Conference on Computer Vision and
  Pattern Recognition}, 11408--11417.

\bibitem[{Dong et~al.(2019)Dong, Yao, Gholami, Mahoney, and
  Keutzer}]{dong2019hawq}
Dong, Z.; Yao, Z.; Gholami, A.; Mahoney, M.; and Keutzer, K. 2019.
\newblock HAWQ: Hessian AWare Quantization of Neural Networks with
  Mixed-Precision.
\newblock \emph{arXiv preprint arXiv:1905.03696} .

\bibitem[{Everingham et~al.(2015)Everingham, Eslami, Van~Gool, Williams, Winn,
  and Zisserman}]{everingham2015pascal}
Everingham, M.; Eslami, S.~A.; Van~Gool, L.; Williams, C.~K.; Winn, J.; and
  Zisserman, A. 2015.
\newblock The pascal visual object classes challenge: A retrospective.
\newblock \emph{International journal of computer vision} 111(1): 98--136.

\bibitem[{Figurnov et~al.(2017)Figurnov, Collins, Zhu, Zhang, Huang, Vetrov,
  and Salakhutdinov}]{figurnov2017spatially}
Figurnov, M.; Collins, M.~D.; Zhu, Y.; Zhang, L.; Huang, J.; Vetrov, D.; and
  Salakhutdinov, R. 2017.
\newblock Spatially adaptive computation time for residual networks.
\newblock In \emph{Proceedings of the IEEE Conference on Computer Vision and
  Pattern Recognition}, 1039--1048.

\bibitem[{Goodfellow, Shlens, and Szegedy(2015)}]{ian2014explaining}
Goodfellow, I.; Shlens, J.; and Szegedy, C. 2015.
\newblock Explaining and Harnessing Adversarial Examples.
\newblock In \emph{International Conference on Learning Representations}.
\newblock \urlprefix\url{http://arxiv.org/abs/1412.6572}.

\bibitem[{Hariharan et~al.(2011)Hariharan, Arbelaez, Bourdev, Maji, and
  Malik}]{BharathICCV2011}
Hariharan, B.; Arbelaez, P.; Bourdev, L.; Maji, S.; and Malik, J. 2011.
\newblock Semantic Contours from Inverse Detectors.
\newblock In \emph{International Conference on Computer Vision (ICCV)}.

\bibitem[{He et~al.(2015)He, Zhang, Ren, and Sun}]{he2015delving}
He, K.; Zhang, X.; Ren, S.; and Sun, J. 2015.
\newblock Delving deep into rectifiers: Surpassing human-level performance on
  imagenet classification.
\newblock In \emph{Proceedings of the IEEE international conference on computer
  vision}, 1026--1034.

\bibitem[{He et~al.(2016)He, Zhang, Ren, and Sun}]{he2016deep}
He, K.; Zhang, X.; Ren, S.; and Sun, J. 2016.
\newblock Deep residual learning for image recognition.
\newblock In \emph{Proceedings of the IEEE conference on computer vision and
  pattern recognition}, 770--778.

\bibitem[{Hinton(2012)}]{hinton2012ste}
Hinton, G. 2012.
\newblock Neural Networks for Machine Learning.

\bibitem[{Hinton, Vinyals, and Dean(2015)}]{hinton2015distilling}
Hinton, G.; Vinyals, O.; and Dean, J. 2015.
\newblock Distilling the knowledge in a neural network.
\newblock \emph{arXiv preprint arXiv:1503.02531} .

\bibitem[{Howard et~al.(2019)Howard, Sandler, Chu, Chen, Chen, Tan, Wang, Zhu,
  Pang, Vasudevan et~al.}]{howard2019searching}
Howard, A.; Sandler, M.; Chu, G.; Chen, L.-C.; Chen, B.; Tan, M.; Wang, W.;
  Zhu, Y.; Pang, R.; Vasudevan, V.; et~al. 2019.
\newblock Searching for mobilenetv3.
\newblock \emph{arXiv preprint arXiv:1905.02244} .

\bibitem[{Ioffe and Szegedy(2015)}]{ioffe2015batch}
Ioffe, S.; and Szegedy, C. 2015.
\newblock Batch normalization: Accelerating deep network training by reducing
  internal covariate shift.
\newblock \emph{arXiv preprint arXiv:1502.03167} .

\bibitem[{Jung et~al.(2019)Jung, Son, Lee, Son, Han, Kwak, Hwang, and
  Choi}]{jung2019learning}
Jung, S.; Son, C.; Lee, S.; Son, J.; Han, J.-J.; Kwak, Y.; Hwang, S.~J.; and
  Choi, C. 2019.
\newblock Learning to quantize deep networks by optimizing quantization
  intervals with task loss.
\newblock In \emph{Proceedings of the IEEE Conference on Computer Vision and
  Pattern Recognition}, 4350--4359.

\bibitem[{Kingma and Ba(2014)}]{kingma2014adam}
Kingma, D.~P.; and Ba, J. 2014.
\newblock Adam: A method for stochastic optimization.
\newblock \emph{arXiv preprint arXiv:1412.6980} .

\bibitem[{Krizhevsky, Hinton et~al.(2009)}]{krizhevsky2009learning}
Krizhevsky, A.; Hinton, G.; et~al. 2009.
\newblock Learning multiple layers of features from tiny images.
\newblock Technical report, Citeseer.

\bibitem[{Krizhevsky, Sutskever, and Hinton(2012)}]{krizhevsky2012imagenet}
Krizhevsky, A.; Sutskever, I.; and Hinton, G.~E. 2012.
\newblock Imagenet classification with deep convolutional neural networks.
\newblock In \emph{Advances in neural information processing systems},
  1097--1105.

\bibitem[{Liu et~al.(2018{\natexlab{a}})Liu, Zoph, Neumann, Shlens, Hua, Li,
  Fei-Fei, Yuille, Huang, and Murphy}]{liu2018progressive}
Liu, C.; Zoph, B.; Neumann, M.; Shlens, J.; Hua, W.; Li, L.-J.; Fei-Fei, L.;
  Yuille, A.; Huang, J.; and Murphy, K. 2018{\natexlab{a}}.
\newblock Progressive neural architecture search.
\newblock In \emph{Proceedings of the European Conference on Computer Vision
  (ECCV)}, 19--34.

\bibitem[{Liu et~al.(2018{\natexlab{b}})Liu, Wu, Luo, Yang, Liu, and
  Cheng}]{liu2018bi}
Liu, Z.; Wu, B.; Luo, W.; Yang, X.; Liu, W.; and Cheng, K.-T.
  2018{\natexlab{b}}.
\newblock Bi-real net: Enhancing the performance of 1-bit cnns with improved
  representational capability and advanced training algorithm.
\newblock In \emph{Proceedings of the European Conference on Computer Vision
  (ECCV)}, 722--737.

\bibitem[{Nagel et~al.(2019)Nagel, van Baalen, Blankevoort, and
  Welling}]{nagel2019data}
Nagel, M.; van Baalen, M.; Blankevoort, T.; and Welling, M. 2019.
\newblock Data-Free Quantization through Weight Equalization and Bias
  Correction.
\newblock \emph{arXiv preprint arXiv:1906.04721} .

\bibitem[{Netzer et~al.(2011)Netzer, Wang, Coates, Bissacco, Wu, and
  Ng}]{netzer2011reading}
Netzer, Y.; Wang, T.; Coates, A.; Bissacco, A.; Wu, B.; and Ng, A.~Y. 2011.
\newblock Reading digits in natural images with unsupervised feature learning .

\bibitem[{Paszke et~al.(2017)Paszke, Gross, Chintala, Chanan, Yang, DeVito,
  Lin, Desmaison, Antiga, and Lerer}]{paszke2017automatic}
Paszke, A.; Gross, S.; Chintala, S.; Chanan, G.; Yang, E.; DeVito, Z.; Lin, Z.;
  Desmaison, A.; Antiga, L.; and Lerer, A. 2017.
\newblock Automatic Differentiation in {PyTorch}.
\newblock In \emph{NIPS Autodiff Workshop}.

\bibitem[{Phan et~al.(2020)Phan, He, Savvides, Shen et~al.}]{phan2020mobinet}
Phan, H.; He, Y.; Savvides, M.; Shen, Z.; et~al. 2020.
\newblock Mobinet: A mobile binary network for image classification.
\newblock In \emph{The IEEE Winter Conference on Applications of Computer
  Vision}, 3453--3462.

\bibitem[{Rastegari et~al.(2016)Rastegari, Ordonez, Redmon, and
  Farhadi}]{rastegari2016xnor}
Rastegari, M.; Ordonez, V.; Redmon, J.; and Farhadi, A. 2016.
\newblock Xnor-net: Imagenet classification using binary convolutional neural
  networks.
\newblock In \emph{European Conference on Computer Vision}, 525--542. Springer.

\bibitem[{Sandler et~al.(2018)Sandler, Howard, Zhu, Zhmoginov, and
  Chen}]{sandler2018mobilenetv2}
Sandler, M.; Howard, A.; Zhu, M.; Zhmoginov, A.; and Chen, L.-C. 2018.
\newblock Mobilenetv2: Inverted residuals and linear bottlenecks.
\newblock In \emph{Proceedings of the IEEE conference on computer vision and
  pattern recognition}, 4510--4520.

\bibitem[{Tan and Le(2019)}]{tan2019efficientnet}
Tan, M.; and Le, Q.~V. 2019.
\newblock EfficientNet: Rethinking Model Scaling for Convolutional Neural
  Networks.
\newblock \emph{arXiv preprint arXiv:1905.11946} .

\bibitem[{Tang, Hua, and Wang(2017)}]{tang2017train}
Tang, W.; Hua, G.; and Wang, L. 2017.
\newblock How to train a compact binary neural network with high accuracy?
\newblock In \emph{Thirty-First AAAI Conference on Artificial Intelligence}.

\bibitem[{Teerapittayanon, McDanel, and
  Kung(2016)}]{teerapittayanon2016branchynet}
Teerapittayanon, S.; McDanel, B.; and Kung, H.-T. 2016.
\newblock Branchynet: Fast inference via early exiting from deep neural
  networks.
\newblock In \emph{2016 23rd International Conference on Pattern Recognition
  (ICPR)}, 2464--2469. IEEE.

\bibitem[{Veit and Belongie(2018)}]{veit2018convolutional}
Veit, A.; and Belongie, S. 2018.
\newblock Convolutional networks with adaptive inference graphs.
\newblock In \emph{Proceedings of the European Conference on Computer Vision
  (ECCV)}, 3--18.

\bibitem[{Wang et~al.(2019)Wang, Liu, Lin, Lin, and Han}]{wang2019haq}
Wang, K.; Liu, Z.; Lin, Y.; Lin, J.; and Han, S. 2019.
\newblock HAQ: Hardware-Aware Automated Quantization with Mixed Precision.
\newblock In \emph{Proceedings of the IEEE Conference on Computer Vision and
  Pattern Recognition}, 8612--8620.

\bibitem[{Wu et~al.(2018)Wu, Nagarajan, Kumar, Rennie, Davis, Grauman, and
  Feris}]{wu2018blockdrop}
Wu, Z.; Nagarajan, T.; Kumar, A.; Rennie, S.; Davis, L.~S.; Grauman, K.; and
  Feris, R. 2018.
\newblock Blockdrop: Dynamic inference paths in residual networks.
\newblock In \emph{Proceedings of the IEEE Conference on Computer Vision and
  Pattern Recognition}, 8817--8826.

\bibitem[{Yu and Huang(2019)}]{yu2019universally}
Yu, J.; and Huang, T. 2019.
\newblock Universally slimmable networks and improved training techniques.
\newblock \emph{arXiv preprint arXiv:1903.05134} .

\bibitem[{Yu et~al.(2018)Yu, Yang, Xu, Yang, and Huang}]{yu2018slimmable}
Yu, J.; Yang, L.; Xu, N.; Yang, J.; and Huang, T. 2018.
\newblock Slimmable neural networks.
\newblock \emph{arXiv preprint arXiv:1812.08928} .

\bibitem[{Zhang et~al.(2018)Zhang, Yang, Ye, and Hua}]{zhang2018lq}
Zhang, D.; Yang, J.; Ye, D.; and Hua, G. 2018.
\newblock Lq-nets: Learned quantization for highly accurate and compact deep
  neural networks.
\newblock In \emph{Proceedings of the European Conference on Computer Vision
  (ECCV)}, 365--382.

\bibitem[{Zhao et~al.(2018)Zhao, Li, Shen, Liang, and Wu}]{zhao2018modulation}
Zhao, X.; Li, H.; Shen, X.; Liang, X.; and Wu, Y. 2018.
\newblock A modulation module for multi-task learning with applications in
  image retrieval.
\newblock In \emph{Proceedings of the European Conference on Computer Vision
  (ECCV)}, 401--416.

\bibitem[{Zhou et~al.(2016)Zhou, Wu, Ni, Zhou, Wen, and Zou}]{zhou2016dorefa}
Zhou, S.; Wu, Y.; Ni, Z.; Zhou, X.; Wen, H.; and Zou, Y. 2016.
\newblock Dorefa-net: Training low bitwidth convolutional neural networks with
  low bitwidth gradients.
\newblock \emph{arXiv preprint arXiv:1606.06160} .

\bibitem[{Zhuang et~al.(2018)Zhuang, Shen, Tan, Liu, and
  Reid}]{zhuang2018towards}
Zhuang, B.; Shen, C.; Tan, M.; Liu, L.; and Reid, I. 2018.
\newblock Towards effective low-bitwidth convolutional neural networks.
\newblock In \emph{Proceedings of the IEEE Conference on Computer Vision and
  Pattern Recognition}, 7920--7928.

\end{thebibliography}
